\def\year{2022}\relax
\documentclass[letterpaper]{article} % DO NOT CHANGE THIS
\usepackage{aaai22}  % DO NOT CHANGE THIS
\usepackage{times}  % DO NOT CHANGE THIS
\usepackage{helvet}  % DO NOT CHANGE THIS
\usepackage{courier}  % DO NOT CHANGE THIS
\usepackage[hyphens]{url}  % DO NOT CHANGE THIS
\usepackage{graphicx} % DO NOT CHANGE THIS
\usepackage{natbib}  % DO NOT CHANGE THIS AND DO NOT ADD ANY OPTIONS TO IT
\usepackage{caption} % DO NOT CHANGE THIS AND DO NOT ADD ANY OPTIONS TO IT
\DeclareCaptionStyle{ruled}{labelfont=normalfont,labelsep=colon,strut=off} % DO NOT CHANGE THIS
\usepackage{algorithm}
\usepackage{algorithmic}

\usepackage{amssymb} % zi ji jia de!
\usepackage{array} %调整表格线与上下内容的间隔
\newcolumntype{P}[1]{>{\centering\arraybackslash}p{#1}} %表格格式定义
\usepackage{booktabs} %调整表格线与上下内容的间隔
\usepackage{multirow} %调整表格线与上下内容的间隔

\usepackage{graphicx}
\usepackage{float}
\usepackage{subfig}

\usepackage[normalem]{ulem}
\useunder{\uline}{\ul}{}

\usepackage{newfloat}
\usepackage{listings}
\floatstyle{ruled}
\newfloat{listing}{tb}{lst}{}
\floatname{listing}{Listing}
\title{Block Modeling-Guided Graph Convolutional Neural Networks}
\author{
    Dongxiao He$^{1}$, Chundong Liang$^{1}$, Huixin Liu$^{1}$, Mingxiang Wen$^{1}$, Pengfei Jiao$^{2,}$\thanks{ Corresponding authors.}, Zhiyong Feng$^{1,*}$
}
\affiliations{
    $^{1}$College of Intelligence and Computing, Tianjin University, Tianjin, China\\
    $^{2}$School of Cyberspace, Hangzhou Dianzi University, Hangzhou, China\\
    \{hedongxiao, liangchundong, liuhuixin, zyfeng\}@tju.edu.cn, ynwmx@qq.com, cspjiao@gmail.com
}

\begin{document}
 \maketitle

\begin{abstract}
Graph Convolutional Network (GCN) has shown remarkable potential of exploring graph representation. However, the GCN aggregating mechanism fails to generalize to networks with heterophily where most nodes have neighbors from different classes, which commonly exists in real-world networks. In order to make the propagation and aggregation mechanism of GCN suitable for both homophily and heterophily (or even their mixture), we introduce block modeling into the framework of GCN so that it can realize \emph{
“block-guided classified aggregation”}, and automatically learn the corresponding aggregation rules for neighbors of different classes. By incorporating block modeling into the aggregation process, GCN is able to aggregate information from homophilic and heterophilic neighbors discriminately according to their homophily degree. We compared our algorithm with state-of-art methods which deal with the heterophily problem. Empirical results demonstrate the superiority of our new 
approach over existing methods in heterophilic datasets while maintaining a competitive performance in homophilic datasets.
\end{abstract}

\section{Introduction}
In recent years, graph-based information exploration has been extensively studied in deep learning~\cite{gnnsurvey1,gnnsurvey2,graphembsurvey1, graphembsurvey2,jin-survey}. Graph Convolutional Network (GCN)~\cite{gcn}, which is famous for its effectiveness in graph representation learning, follows the principal learning mechanism where adjacency nodes attain similar representations through aggregating their neighboring information. The representations support subsequent downstream tasks such as node classification~\cite{graphclassify1,hgnn-ac,ArmGAN,nodeClassification1} and link prediction~\cite{linkprediction1,linkprediction2,linkprediction3,cao2021dual}.

Despite its wide application, GCN and its variants~\cite{sgc,fastgcn} are typically limited by the implicit homophily assumption where nodes within the proximity have similar representations, that is, birds of a feather flock together~\cite{homo-social}. But this is not satisfied in many real-world heterophilic datasets, e.g., fraud detection networks or the protein structure graphs, where nodes from different classes tend to make connections due to opposites attract. Some recent studies show that the performance of GCN can be severely restricted in this type of datasets, due to the fact that GCN's aggregating mechanism is not designed for heterophily settings.

Recently some methods aiming to solve the GCN homophily problems have been proposed. Based on the designing methodologies, these algorithms can be mainly divided into two types. 1) Aggregating higher-order neighborhoods, such as H2GCN~\cite{h2gcn} and MixHop~\cite{mixhop}. These algorithms are based on the idea that direct neighborhoods may be heterophily-dominant, but the higher-order neighborhoods are homophily-dominant and thereby provide more valuable information. Therefore, by explicitly aggregating information from higher-order neighborhood, the heterophily problem of GCN brought by aggregating information from immediate neighborhoods can be alleviated. 2) Passing signed messages between heterophilic neighbors, such as GGCN~\cite{ggcn} and GPR-GNN~\cite{gprgnn}. These algorithms normally assign a weight to every connected node based on the similarity between them. As the nodes aggregate information from neighbors, they get positive messages from neighbors with the same class while negative messages from neighbors with different classes. In this way, positive messages allow neighbors of similar class to intensify their impact, while negative messages prevent dissimilar neighbors from bringing irrelevant information which may harm performance.

However, existing algorithms for heterophily have two main drawbacks. The first is the damage of network topology. Existing methods typically expand high-order nodes as neighbors and try to find more homophilic information, which will change the original topology of networks. But in network science, network topology, even having heterophilic connecting pattern, possess vital information which can be preserved and fully utilized. The second is the limitation of aggregating mechanism. Methods like MixHop and GPR-GNN extend the same treatment to all neighbors in different classes which is not satisfactory, while methods like GGCN is an opposite extreme that gives every neighbor a weight parameter which is highly computational expensive.

Then, an intuitive solution may be that we allow neighbors of the same class aggregated in the same way while neighbors of different classes aggregated in different ways. Block modeling which is to describe structural regularities (including homophily, heterophily, or their mixture) of networks should have a big potential to address this problem. But unfortunately, while block modeling depicts regular relationships between classes via a so called block matrix (which describes the possibility of nodes in two blocks connected by an edge), it still cannot be used for guiding a classified aggregation in the graph convolutional framework.
\begin{enumerate}
    \item The first challenge is how to derive the block matrix in GCN since it is only a statistical modeling for network structural regularities. For getting block matrix, the block class labels of all nodes must be known. GCN is a semi-supervised learning model and only part of labels are available. So in order to solve this problem, we introduce a multilayer perception (MLP) into the whole learning framework and use it to learn soft labels for all nodes using attribute information. And then we use the soft labels to derive the block matrix.
    \item The second and more important challenge is how to derive the classified aggregation mechanism based on the derived block matrix since this matrix (which depicts the probability distribution for generating an edge between nodes in any two blocks) cannot be used for guiding classified aggregation directly. For this problem, we propose to create a new block similarity matrix based on the derived block matrix, which can characterize the similarity degree between different blocks in the connecting pattern of blocks. By doing so, we can use this new matrix as an aggregation indicator to construct the graph convolutional operation and finally realize the classified aggregation for both homophilic and heterophilic graphs. 
\end{enumerate}

Then, based on the above idea, we present a new \textbf{G}raph \textbf{C}onvolutional \textbf{N}etwork with \textbf{B}lock \textbf{M}odelling-guided-classified aggregation, namely BM-GCN, which is suitable for both homophilic and heterophilic situations. In the proposed BM-GCN, the MLP is applied to learn unknown labels and then derive the block similarity matrix, and the derived block similarity matrix as well as the learned labels can co-guide attribute information propagating and aggregating on network topology. The process of learning block similarity matrix and block similarity-guided graph convolutional operation are integrated into a unified framework. In this way, the learning of unknown class labels can help realize classified aggregation, and block-guided graph convolutional operation can further help MLP improve its performance on learning the unknown labels.

\section{Preliminaries}
\textbf{Problem Setup.} A attributed graph can be formulated as $\mathcal{G}=\left( \mathcal{V},\mathcal{E}, X \right)$, where $\mathcal{V}$ is the set of nodes, $\mathcal{E}$ the set of edges and $X$ the node attributes. Each row in $X$ indicates an attribute vector of a node. The edge set can also be represented by an adjacency matrix $A\in {{\{0,1\}}^{n\times n}}$, where $n=\left| \mathcal{V} \right|$. For semi-supervised tasks, nodes in training set ($\mathcal{T}{{}_{\mathcal{V}}}$) have ground truth labels. The labels are formulated as a label matrix $Y\in {{\mathbb{R}}^{n\times c}}$ in which each row is a one-hot label vector, where $c$ is the number of classes.

\textbf{Homophily Ratio.} The homophily ratio~\cite{geomgcn} can measure the overall homophily level in a graph. It counts the ratio of same-class neighbor nodes to the total neighbor nodes in a graph, defined as
\begin{equation}\label{eq1}
h = \frac{1}{{\left| {\cal V} \right|}}\sum\nolimits_{{v_i} \in {\cal V}} {\frac{{\left| {\{ {v_j}|{v_j} \in {{\cal N}_i},{Y_j} = {Y_i}\} } \right|}}{{\left| {{{\cal N}_i}} \right|}}}
\end{equation}
where ${{\cal N}_i}$ is the neighbor set of node ${v_i}$. In this work, we use homophily ratio $h$ to determine whether a graph is homophilic or heterophilic.

\textbf{Block Matrix.} Given the labels $Y\in {{\mathbb{R}}^{n\times c}}$ for all nodes and the adjacency matrix  $A \in {\{ 0,1\} ^{n \times n}}$, the block matrix is defined as
\begin{equation}\label{eq2}
H = \left( {{Y^T}AY} \right)\oslash \left( {{Y^T}AE} \right)
\end{equation}
where $E$ an all-ones matrix with the same size as $Y$, and $\oslash$ the Hadamard (element-wise) division operation. Block matrix models the linked possibility of nodes in any two blocks. In this work, blocks represent the classes of labels in a graph. From the node-wise level, ${{H}_{i,j}}$ is the probability that a node in the $i$-th class connects with a node in the $j$-th class.

The notations are summarized in Table~\ref{tab:notation}.
 
\begin{table}[t]
  \resizebox{0.98\linewidth}{!}{
  \begin{tabular}{l|c}
    \toprule[1.5pt]
    Notations&Explanations\\
    \midrule[0.6pt]
	$\mathcal{G}$ & A graph\\
    $\mathcal{V}$ & The set of nodes in graph $\mathcal{G}$\\
	$\mathcal{E}$ & The set of edges in graph $\mathcal{G}$\\
	$\mathcal{T}_{\mathcal{V}}$ & Training node set\\
	$\mathcal{N}_{i}$ & The set of neighbor of node $v_i$\\
	$A$ & Adjacency matrix\\
	$X$, $X_i$ & Attribute matrix, attribute vector of node $v_i$\\
	$Y$, $Y_i$ & Label matrix, one-hot label vector of node $v_i$\\
	$B$, $B_i$ & Soft label matrix, soft label vector of node $v_i$\\
	$H$, $h_{i,j}$ & Block matrix, an element in $H$\\
	$Q$, $q_{i,j}$ & Block similarity matrix, an element in $Q$\\
	$Z$ & Node representations\\
    \bottomrule[1.5pt]
  \end{tabular}
  }
  \caption{Notations and Explanations.}\label{tab:notation}
\end{table}

\section{Methods}
Here we show the proposed method, starting with an overview, following the detail designs, and finally give illustrative examples to show the effectiveness of the core design.

\begin{figure*}[th]
	\centering
	\includegraphics[width=0.75\linewidth]{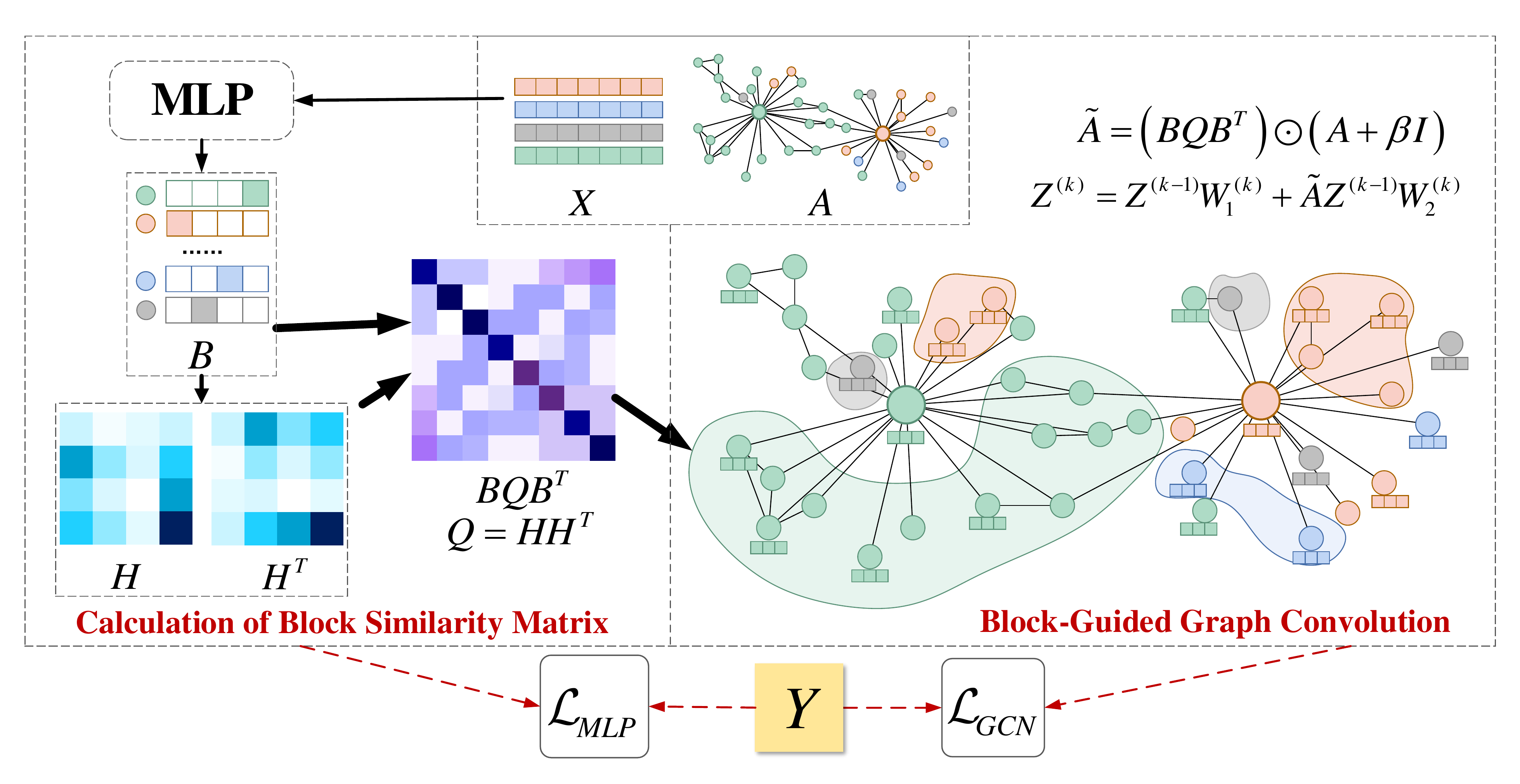}
	\caption{The structure of BM-GCN. It consists of two parts, the calculation of block similarity matrix and the block-guided graph convolution. In the block similarity matrix calculating part, a MLP layer is first applied to generate soft labels $B$, then block matrix $H$ and block similarity matrix $Q$ are computed based on soft labels $B$. In block-guided graph convolution part, the graph convolutional operation is conducted under the guidance of block similarity matrix $Q$ and soft labels $B$. Particularly, BM-GCN achieves a \emph{classified aggregation} mechanism in graph convolutional operation via $BQ{{B}^{T}}$ optimization (different color means different aggregation for different classes). Finally, two semi-supervised loss are combined to optimize BM-GCN model in an end-to-end manner.}
	\label{overview}
\end{figure*}
\subsection{Overview}
To solve the problem that GCN is constrained by homophily assumption, we introduce block modeling into GCN and developed a new graph convolutional network that is suitable for both homophilic and heterophilic situations. This new framework contains two parts: one is to learn the block matrix and the other is to derive a new mechanism based on block matrix and use it to conduct new propagation and aggregation (which are key operations in GCN). For learning block matrix, we need labels of all nodes, while GCN is typically a semi-supervised method~\cite{Lap}, i.e., only part of labels is available. For solving this problem, we use a multi-layer perception (MLP) to learn soft labels for nodes without class labels. After getting labels of all nodes, we use a combination of the given and learned labels as well as the topological structure of networks to compute the block matrix. Block matrix denotes the probability for generating an edge between nodes in any two blocks (classes). Block matrix reflects homophily and heterophily among parts of networks while it cannot be used directly to conduct propagation and aggregation. So in this case, we developed a new aggregation mechanism, i.e., block-guided classified aggregation, by a smartly using of the block matrix. In specific, we use the block matrix to define a block similarity matrix which measures the similarity degree of any two blocks based on connecting patterns among blocks, and contains homophilic and heterophilic information among blocks. Block similarity matrix provides potentially valuable rules for information propagation in graph convolutional layers, which can ignore the homophily or heterophily of a graph and finally achieve a classified aggregation. Then, we use this block similarity matrix and the learned soft labels from MLP to redefine a new aggregation operation in which class labels of nodes and the similarities among blocks can co-guide the attribute information propagating and aggregating on network topology. At last, these two parts of BM-GCN, the learning of block matrix and the block-guided propagation and aggregation, work together to form a unified graph convolutional framework which is optimized jointly by back propagation. 
\subsection{Learning of Block Matrix}
We first need to learn the block matrix. Indicated by Eq.~(\ref{eq2}), computing block matrix requires complete labels. However, graph convolutional networks are semi-supervised models, in which a large number of labels are unknown. In order to fill this gap, BM-GCN adopts the way of learning the unknown labels~\cite{softlabel} from the given data (the known labels  and the attribute network) to compute the block matrix. Considering that the soft labels should come from the original data while the topology may be not trustworthy in heterophilic graphs, here BM-GCN uses node attributes alone to generate soft labels. Specifically, BM-GCN adopts a multilayer perception (MLP) to transform node attributes into soft labels
\begin{equation}\label{eq3}
  \bar B = \sigma \left( {{\rm{MLP}}\left( X \right)} \right)
\end{equation}
where $X$ is node attributes, $\bar{B}$ the output of MLP, and $\sigma \left( \cdot  \right)$ an activation function. Then a softmax operation are applied to $\bar{B}$ for generating soft labels
\begin{equation}\label{eq4}
    B = {\rm{softmax}}\left(\bar  B \right)
\end{equation}

In order to ensure the reliability of the soft label $B$, BM-GCN first pre-trains the MLP layer with the training ground-truth labels for several iterations. Specifically, the pre-training process aims to minimize the loss function
\begin{equation}\label{eq5}
   {{\cal L}_{MLP}} = \sum\nolimits_{{v_i} \in {{\cal T}_{\cal V}}} {f\left( {{{B}_i},{Y_i}} \right)}  
\end{equation}
where ${{B}_{i}}$ is the soft label of node ${{v}_{i}}$, ${{Y}_{i}}$ is the ground-truth label of ${{v}_{i}}$, ${{\mathcal{T}}_{\mathcal{V}}}$ is the nodes in training set, and $f\left( \cdot  \right)$ is the cross entropy.

After the pre-training process, there are ground-truth labels and soft labels available for nodes in training set, and only soft labels available for the remaining nodes. BM-GCN maximizes the use of available ground-truth labels by assembling the two kinds of labels
\begin{equation}\label{eq6}
    {Y_s} = \{ {Y_i},{B_j}|\forall {v_i} \in {{\cal T}_{\cal V}},\forall {v_j} \notin {{\cal T}_{\cal V}}\} 
\end{equation}
where ${Y_s}$ is the assembled label matrix with each row being an label vector of each node. Then the block matrix can be computed via Eq.~(\ref{eq2}), which can be rewritten as
\begin{equation}\label{eq7}
    H = \left( {Y_s^TA{Y_s}} \right)\oslash \left( {Y_s^TAE} \right)
\end{equation}
where $A$ is the adjacency matrix, $E$ is an all-ones matrix with the same size as ${{Y}_{s}}$, and $\oslash$ is the Hadamard (element-wise) division operation. The block matrix $H$ can represent the connecting pattern between blocks (classes), which can well reflect the homophily ratio of the graph. In particular, the more frequently the nodes with the same soft label are connected, the higher homophily ratio the graph has.

\subsection{Block Similarity Matrix}
The block matrix $H$ describes the possibility distribution of two nodes in any two blocks (classes) to be connected by an edge. However, this matrix cannot directly guide GCN to achieve a classified aggregation process. This is because in a heterophilic graph where edges tend to connect nodes in different classes, the possibility values between different classes may be larger than that within the same class. So, in order to realize the block-guided classified aggregation, it is necessary to modify the element values in block matrix $H$ so that these elements can reflect potentially valuable information on propagating rules between various classes of nodes in the graph convolutional operation. With this purpose, we innovatively propose a new block similarity matrix based on block matrix, which is defined as
\begin{equation}\label{eq8}
    Q = H{H^T}
\end{equation}
The similarity matrix $Q$ measures the similarity degree of different blocks in $H$, indicating that blocks (classes) with similar structural connecting pattern will have more information propagation with each other. Furthermore, since nodes within the same class should have more information exchange, BM-GCN enhance the information propagating ratio within the same class, i.e.,
\begin{equation}\label{eq9}
    {\rm{Diag}}\left( Q \right) \leftarrow \alpha  \cdot {\rm{Diag}}\left( Q \right)
\end{equation}
where ${\rm{Diag}}\left(  \cdot  \right)$ means the diagonal elements of a matrix, and $\alpha $ the enhancement factor.

\subsection{Block-Guided Graph Convolution}
Based on the new created block similarity matrix $Q$, BM-GCN can assign different information propagating rules for different class-combinations. Furthermore, soft labels can indicate which class-combination the two nodes belong to. In this way, the information propagating process can be jointly guided by soft label $B$ and block similarity matrix $Q$. Particularly, consider two nodes ${{v}_{i}}$ and ${{v}_{j}}$ with their soft labels ${{B}_{i}}=\{b_{i}^{1},\ b_{i}^{2},\ ...,\ b_{i}^{c}\}$ and ${{B}_{j}}=\{b_{j}^{1},\ b_{j}^{2},\ ...,\ b_{j}^{c}\}$ respectively, where $c$ is the number of classes. There are ${{c}^{2}}$ candidate class-combinations for node pairs $\left\langle {{v}_{i}},{{v}_{j}} \right\rangle $, each of which can be probabilized as
\begin{equation}\label{eq10}
    p\left( {\varphi ({v_i}) = {Y_r},\;\varphi ({v_j}) = {Y_t}} \right) = b_i^rb_j^t
\end{equation}
where $\varphi \left( \cdot  \right)$ is a function that maps a node to its class, and $r,t\in \{1,2,...,c\}$. Meanwhile, the block similarity matrix $Q$ indicates information propagating probability between any two classes, i.e., the more similar two classes, the more information should be propagated. Therefore, propagating probability between nodes ${{v}_{i}}$ and ${{v}_{j}}$ can be seen as the expectation of elements in $Q$, i.e.,
\begin{equation}\label{eq11}
   {\omega _{ij}} = \sum\limits_{r = 1}^c {\sum\limits_{t = 1}^c {{q_{r,t}}b_i^rb_j^t} } 
\end{equation}
where ${{q}_{r,t}}$ is an element in $Q$, indicating information propagating probabicity between the $r$-th class and the $t$-th class. Eq.~(\ref{eq11}) demonstrates that the propagation probability between nodes ${{v}_{i}}$ and ${{v}_{j}}$ are guided by their soft labels and block similarity matrix $Q$ simultaneously. Then, for all node pairs in a graph, propagating probabilities along with these pairs can be formulated as a weight matrix
\begin{equation}\label{eq12}
    \Omega  = BQ{B^T}
\end{equation}
Eq.~(\ref{eq12}) is the matrix-level expression of Eq.~(\ref{eq11}). Then, we use weight matrix $\Omega $ to refine the topology
\begin{equation}\label{eq13}
    A' = \Omega  \odot \left( {A + \beta I} \right)
\end{equation}
where $I$ is the identity matrix, $\beta $ a hyper-parameter, and $ \odot $ the Hadamard (element-wise) multiplication operation. Then, BM-GCN normalizes the weights on edges by a softmax operation
\begin{equation}\label{eq14}
    {\tilde a_{i,j}} = \frac{{\exp \left( {{{a'}_{i,j}}} \right)}}{{\sum\nolimits_{{v_s} \in {\cal N}} {\exp \left( {{{a'}_{i,s}}} \right)} }}
\end{equation}
where ${{{a}'}_{i,j}}$ is an element in ${A}'$. Here we name the normalized topological matrix as $\tilde{A}$, and replace the normalized graph laplacian used in GCN with our new $\tilde{A}$. In this way, graph convolutional operation in BM-GCN can finally realize a classified aggregation mechanism because the information propagation process is under the guidance of soft labels and block similarity matrix $Q$. Node pairs belonging to different soft label combinations will have different information exchange, and the information exchange ratio is determined by $Q$. Then, the new graph convolutional layer can be written as
\begin{equation}\label{eq15}
    {Z^{(k)}} = {Z^{(k - 1)}}W_1^{(k)} + \tilde A{Z^{(k - 1)}}W_2^{(k)}
\end{equation}
where ${{Z}^{(k)}}$ denotes node representations in the $k$-th layer, $W_{1}^{(k)}$ and $W_{2}^{(k)}$ are learnable parameters specific to the $k$-th layer, and ${{Z}^{(0)}}=X$.

\subsection{Model Optimization}
Similar to the MLP layer (Eq.~\ref{eq5}), BM-GCN adopts a semi-supervised loss as
\begin{equation}\label{eq16}
    {{\mathcal{L}}_{GCN}}=\sum\nolimits_{{{v}_{i}}\in {{\mathcal{T}}_{\mathcal{V}}}}{f\left( {{Z}_{i}},{{Y}_{i}} \right)}
\end{equation}
where $Z$ is the output of the last graph convolutional layer. Here, BM-GCN also optimizes pre-trained MLP layer in a fine-tune manner~\cite{pretrain}. Specifically, BM-GCN integrates the process of block similarity learning and the process of block guided graph convolutional operation into a unified framework. Incorporating the loss function in MLP layer and in graph convolutional operation, the final loss function can be written as
\begin{equation}\label{eq17}
    {{\cal L}_{final}} = \lambda {{\cal L}_{GCN}} + \left( {1 - \lambda } \right){{\cal L}_{MLP}}
\end{equation}
where $\lambda$ is the balance parameter (with the default value 0.5). By minimize ${{\mathcal{L}}_{final}}$, BM-GCN trains all modules in an end-to-end manner.

\begin{figure}[t]  
    \centering    
    \subfloat[] %The ground truth of block matrix $H$
    {
        \begin{minipage}[t]{0.45\textwidth}
            \centering          
            \includegraphics[width=0.45\textwidth]{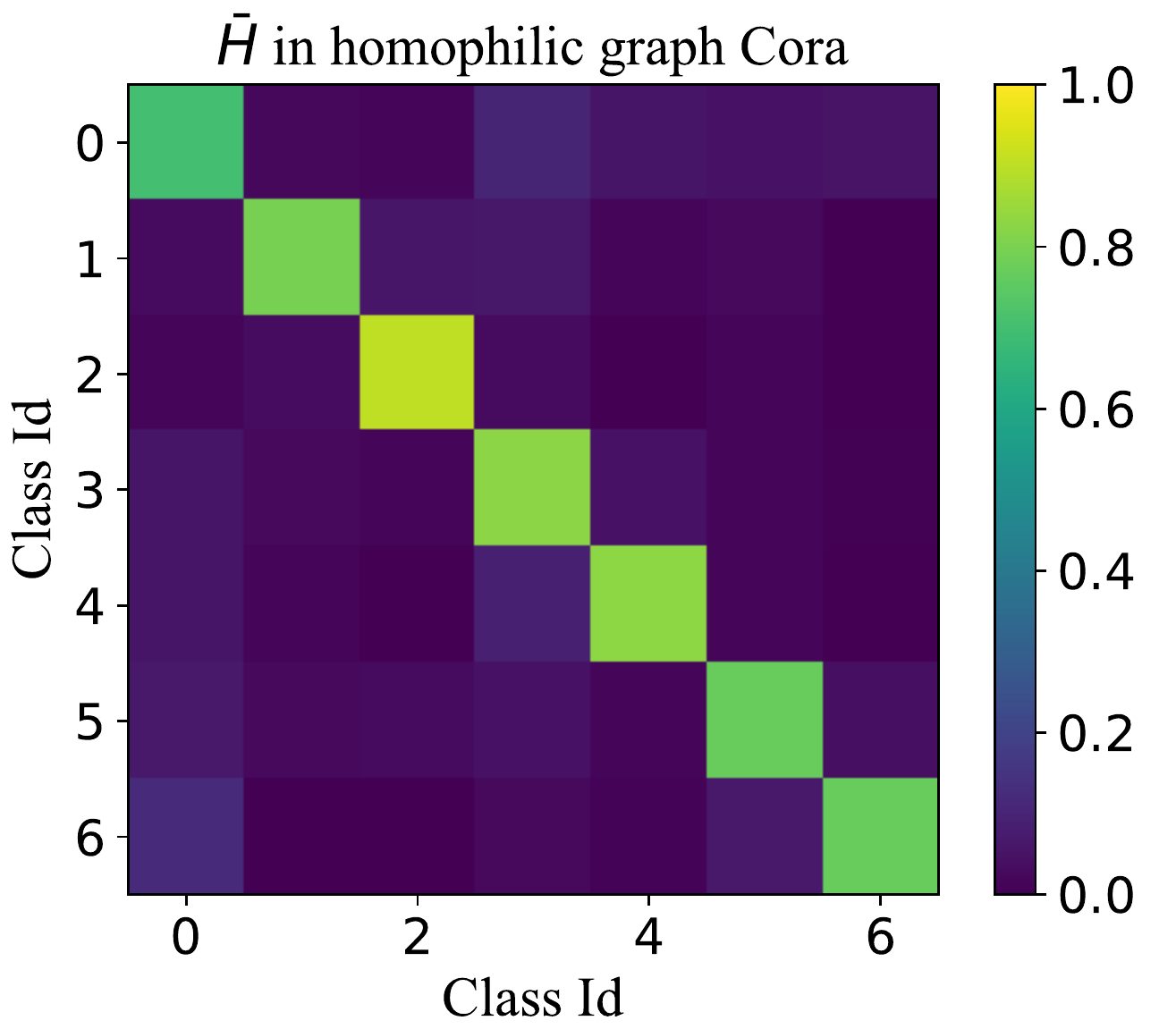}%\hspace{5mm}
            \includegraphics[width=0.45\textwidth]{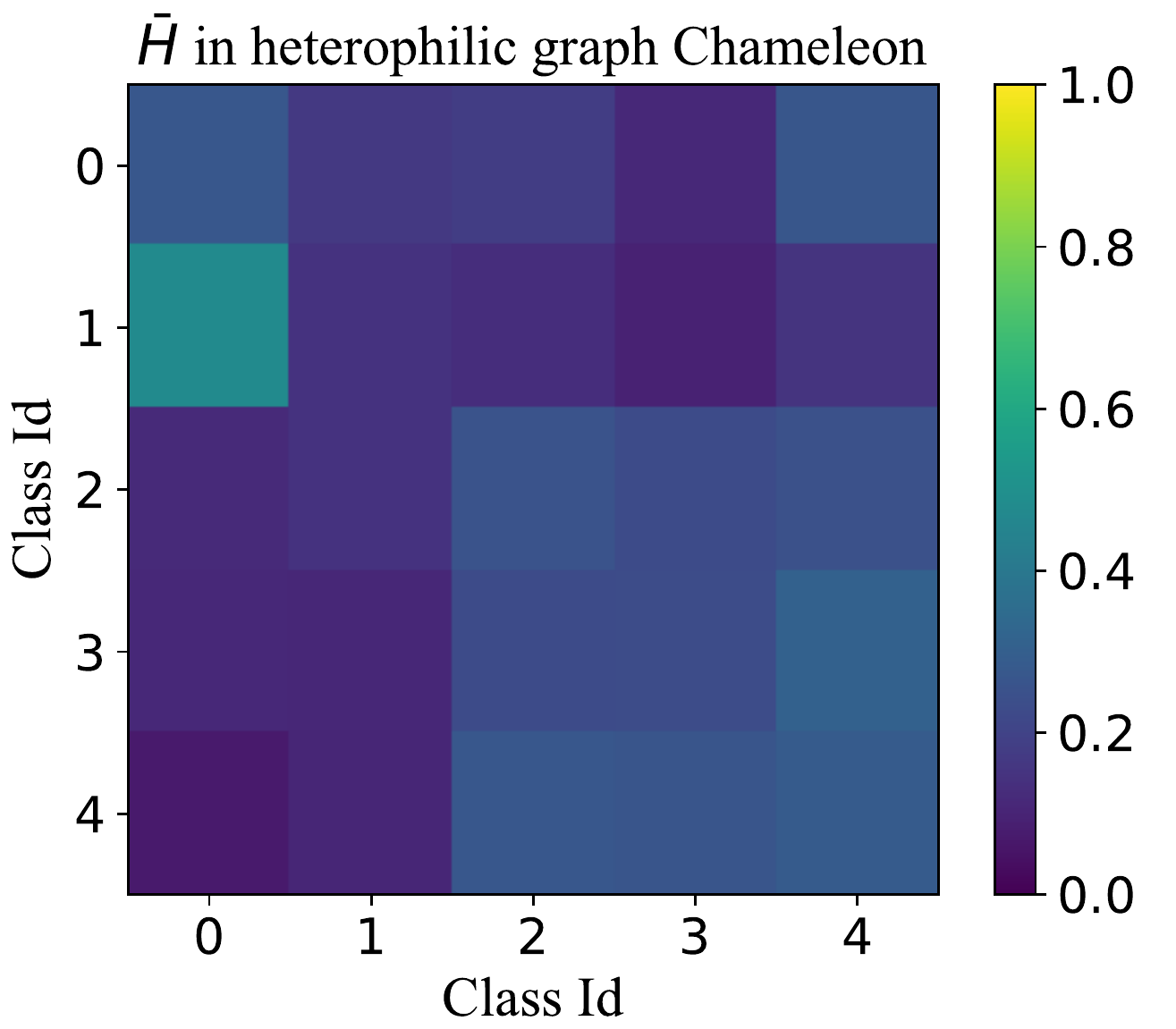}
        \end{minipage}%
    }
    \vspace{0mm}
    \subfloat[] %The learned block matrix $H$
    {
        \begin{minipage}[t]{0.45\textwidth}
            \centering      
            \includegraphics[width=0.45\textwidth]{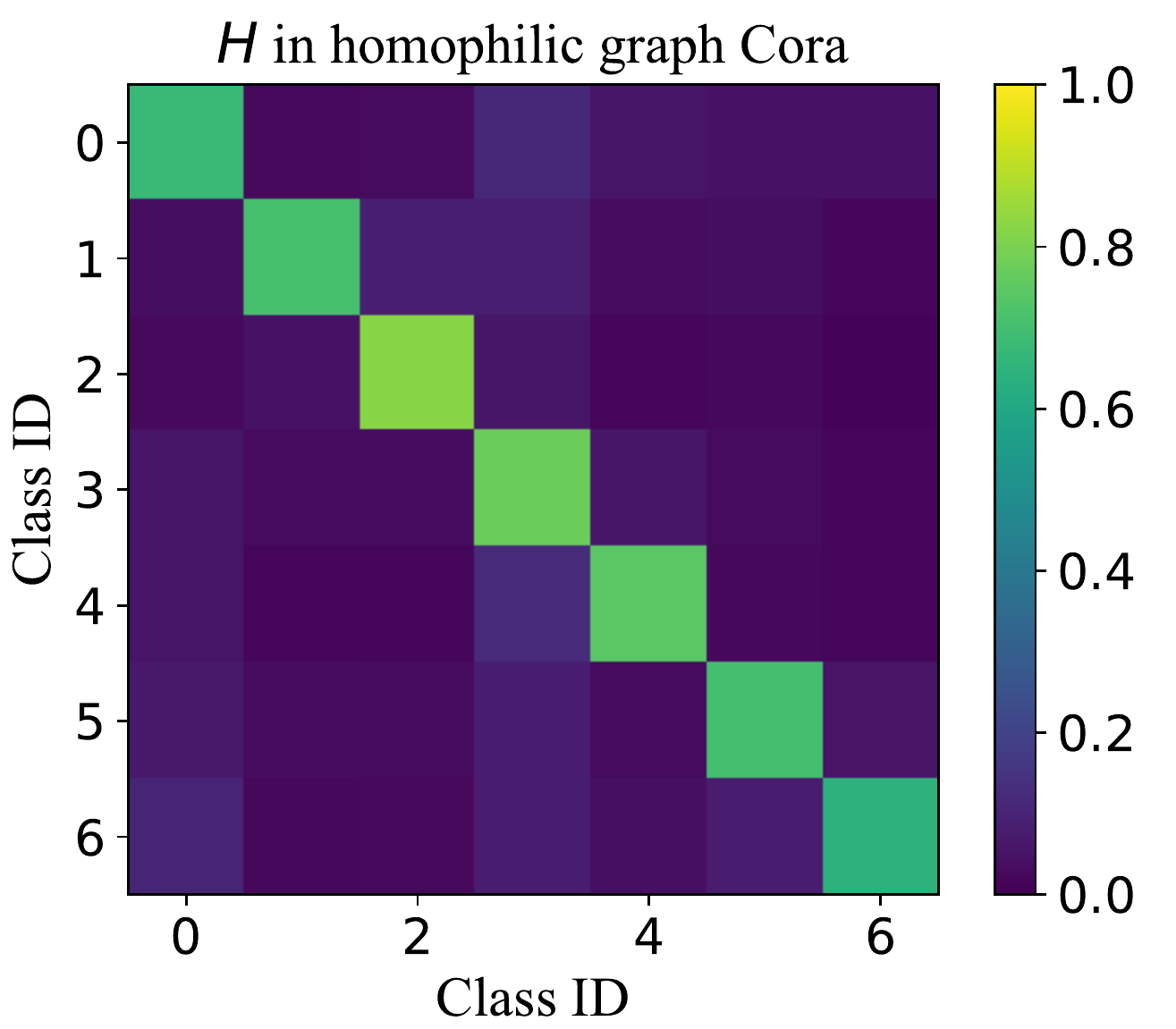}%\hspace{5mm}
            \includegraphics[width=0.45\textwidth]{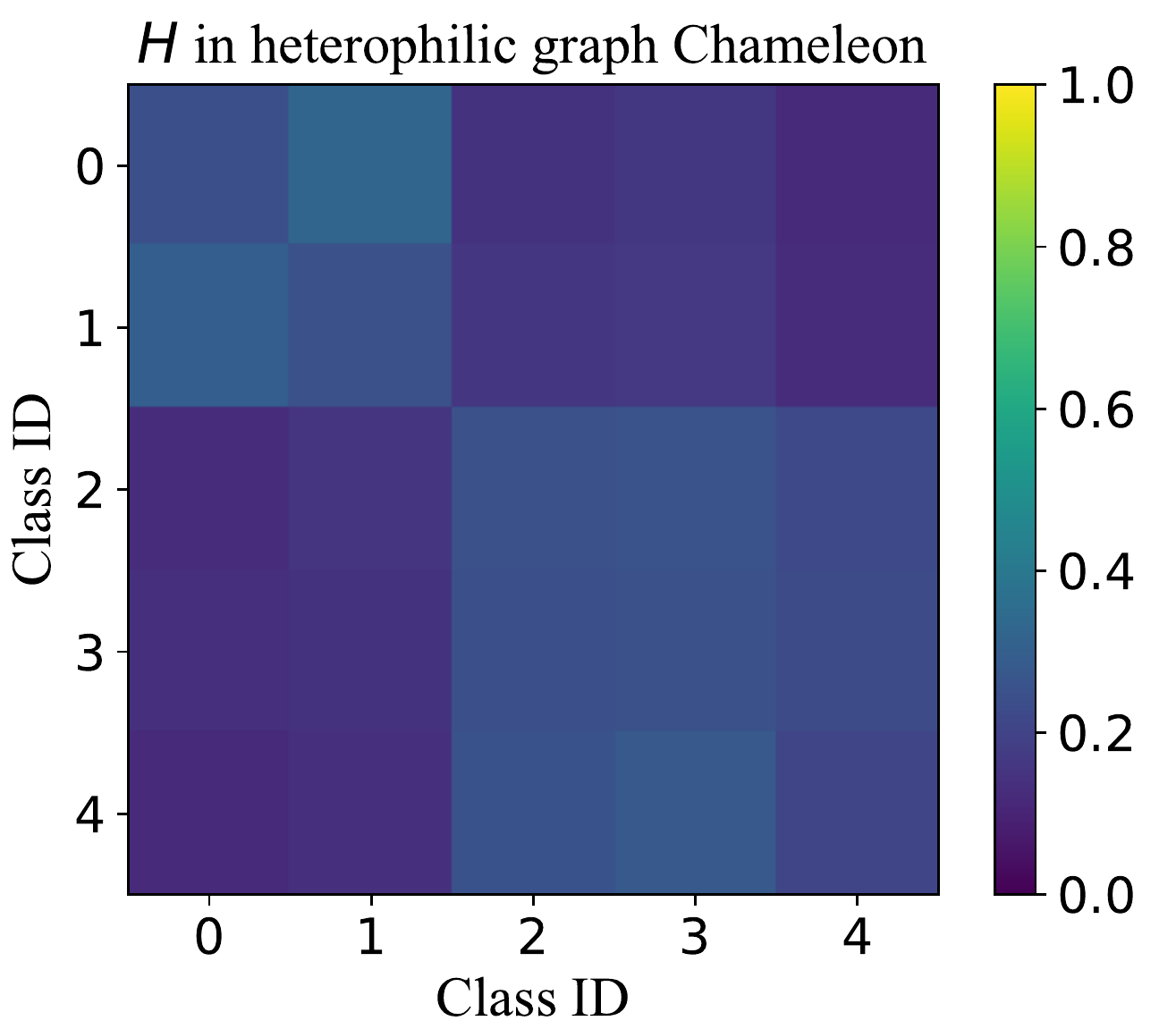}
        \end{minipage}
    }%
    \vspace{0mm}
    \subfloat[] %The learned block similarity matrix $Q$
    {
        \begin{minipage}[t]{0.45\textwidth}
            \centering          
            \includegraphics[width=0.45\textwidth]{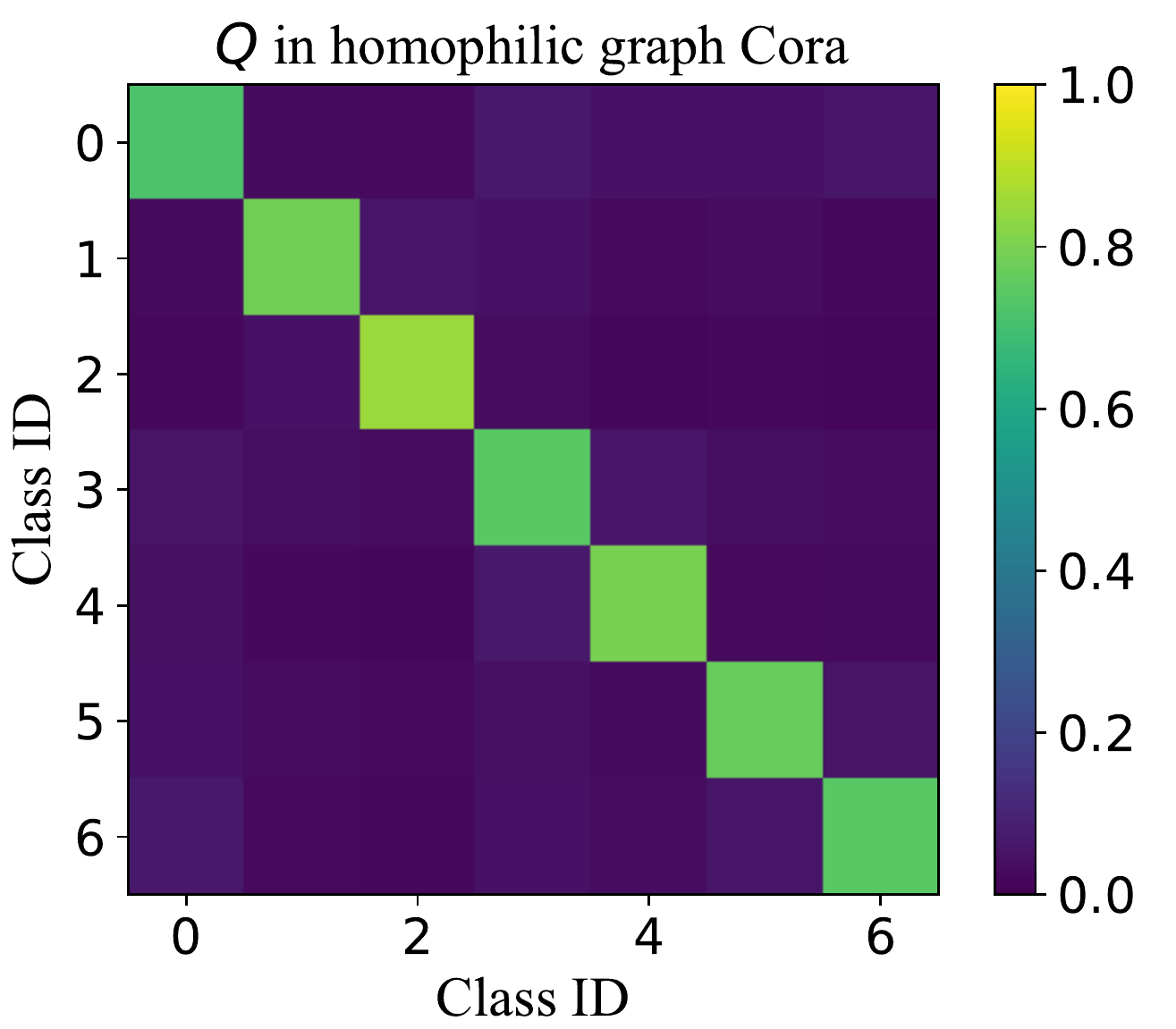}%\hspace{5mm}
            \includegraphics[width=0.45\textwidth]{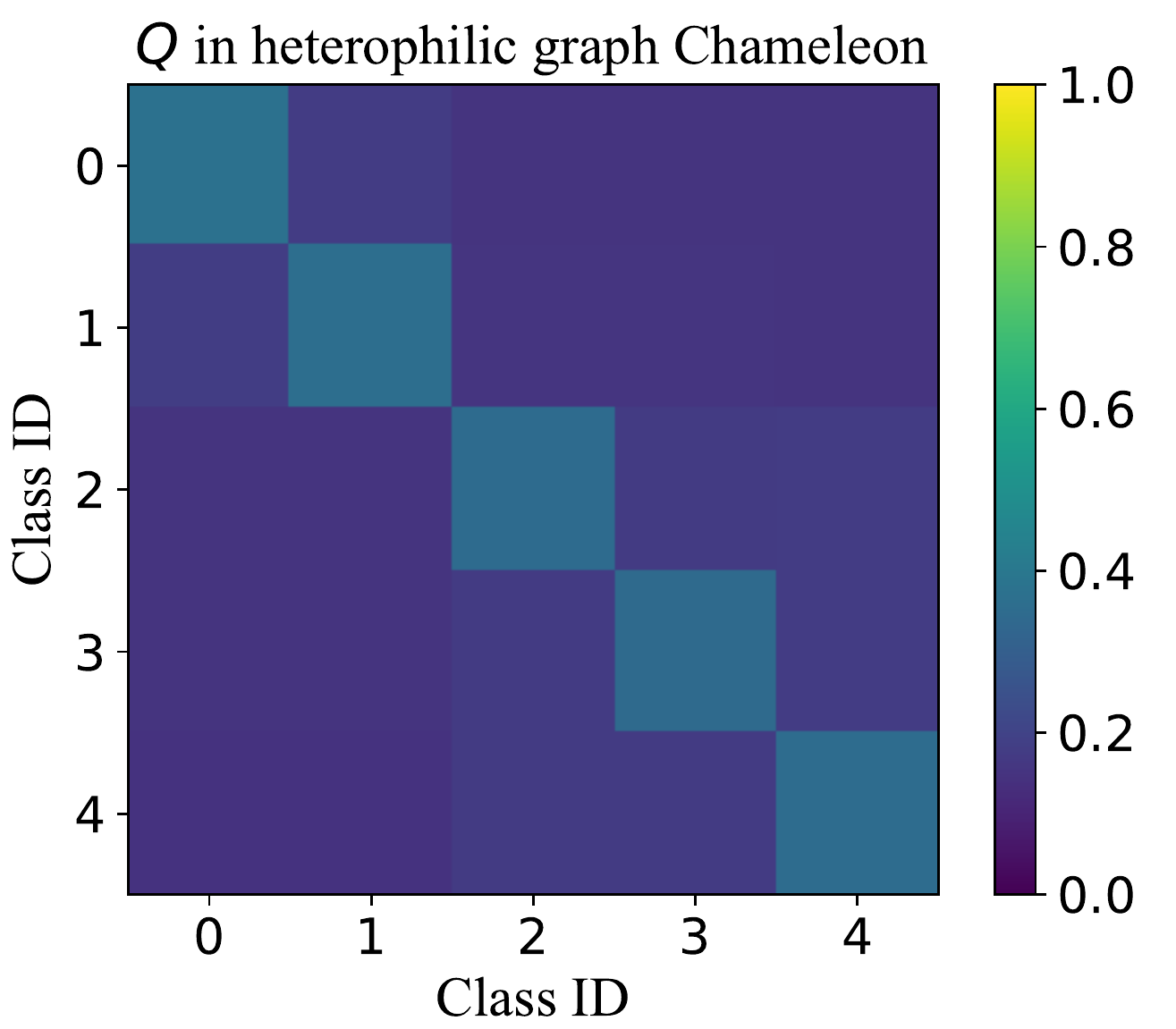}
        \end{minipage}%
    }
    \caption{Visualization of (a) block matrix $\bar H$ calculated based on ground truth, (b) block matrix $H$ learned by our approach, and (c) the new block similarity matrix $Q$ created based on $H$. The left column is the result on homophilic graph Cora and the right is that on heterophilic graph Chameleon. x-axis (and y-axis) denotes the id of the block (class). The lager the brighter for elements in each matrix.}\label{fig:effectiveness}
\end{figure}

\subsection{Illustration on Why Block Modeling Effective}
The proposed BM-GCN aims to realize a block-guided classified aggregation so that nodes sharing the same or similar classes will have more information exchange. In order to achieve this goal, block matrix is introduced into BM-GCN to model the relationships between classes. The elements in the block matrix $H$ indicates the connecting possibility between various classes of nodes. Here we take a homophilic graph Cora~\cite{cora1} and a heterophilic graph Chameleon~\cite{chameleon} as examples to show how BM-GCN works. 

Fig.~\ref{fig:effectiveness}(a) show the block matrix $\bar{H}$ of Cora (or Chameleon) calculated based on the ground truth. As shown, the distribution of these two matrices have different patterns. The connecting possibility of nodes within the same class is larger in the homophilic graph Cora, while the connecting possibility of nodes between different classes is larger in the heterophilic graph Chameleon. Fig.~\ref{fig:effectiveness}(b) shows the block matrix $H$ of Cora (or Chameleon) learned based on the soft labels calculated via Eq.~(\ref{eq4}). As shown, our $H$ is always close to $\bar{H}$ on either Cora or Chameleon, which partly shows the strong learning ability of our new approach.

However, in this case the learned block matrix $H$ can only help aggregate more same-class information on homophilic graphs rather than heterophilic graphs. That means, the off-diagonal elements with big values in $H$ of the heterophilic graph (right in Fig.~\ref{fig:effectiveness}b) will still cause nodes receiving too many noises during graph convolution and thus lead to performance degradation. But fortunately, we create a new block similarity matrix $Q$ based on $H$ using Eq.~(\ref{eq8}), which can well measure the relationship between two classes from a new perspective. That is, \emph{if two classes have the similar connected value distributions with all kinds of classes in a graph, the two classes should be more similar.} The visualizations of $Q$ are shown in Fig.~\ref{fig:effectiveness}(c). As shown, the values of diagonal elements in this new matrix $Q$ are then always bigger than off-diagonal elements, on both homophilic and heterophilic graphs. It successfully achieved the role of block-guided classified aggregation on heterophilic graphs (right in Fig.~\ref{fig:effectiveness}c), and meanwhile, preserved the original distributions of block matrix on homophilic graphs (left in Fig.~\ref{fig:effectiveness}c), ensuring the stable performance on both homophilic and heterophilic graphs.

\section{Experiments}
We first discuss the experiment setup, including datasets, baselines, and parameter settings. We then evaluate the proposed methods on node classification and visualization tasks, and finally give the parameter analysis.

\begin{table}[t]
	\centering
	\resizebox{0.90\linewidth}{!}{
	\begin{tabular}{l|lllll}
        \toprule[1.5pt]
        \textbf{Datasets}  & $\mathcal{|V|}$ & $\mathcal{|E|}$ & $c$ & $d$   & $h$  \\ \midrule[0.6pt]
        \textbf{texas}     & 183           & 295           & 5   & 1703  & 0.11 \\
        \textbf{squirrel}  & 5,201         & 198,493       & 5   & 2,089 & 0.22 \\
        \textbf{chameleon} & 2,277         & 31,421        & 5   & 2.325 & 0.23 \\
        \textbf{cora}      & 3,327         & 4,676         & 7   & 3,703 & 0.74 \\
        \textbf{pubmed}  & 19,717        & 44,327        & 6   & 500   & 0.80 \\
        \textbf{citeseer}    & 2,708         & 5,278         & 3   & 1,433 & 0.81 \\ \bottomrule[1.5pt]
    \end{tabular}
    }
	\caption{Statistics of datasets. $\mathcal{|V|}$, $\mathcal{|E|}$, $c$, $d$, $h$ are the number of nodes, edges, classes, and features, and homophily ratio, respectively.}
	\label{table:datasets}
\end{table}

\begin{table*}[t]
\centering
\resizebox{0.83\linewidth}{!}{
\begin{tabular}{lcccccc}
\toprule[1.5pt]
\multicolumn{1}{l}{\multirow{3}{*}{\textbf{\begin{tabular}[c]{@{}l@{}}Method/\\ Accuracy (\%)\end{tabular}}}} & \multicolumn{3}{c}{\textbf{heterophilic graphs}}                             & \multicolumn{3}{c}{\textbf{homophilic graphs}}                              \\ \cmidrule(r){2-4} \cmidrule(r){5-7} 
\multicolumn{1}{c}{}                                  & \textbf{Texas}          & \textbf{Squirrel}       & \textbf{Chameleon}       & \textbf{Cora}           & \textbf{Citeseer}       & \textbf{Pubmed}         \\
\multicolumn{1}{c}{}                                  & $h=0.09$                & $h=0.23$                & $h=0.22$                 & $h=0.81$                & $h=0.74$                & $h=0.8$                 \\ \midrule[0.6pt]
MLP                                                   & 82.70±6.19          & 33.35±1.24          & 48.20±2.63           & 74.14±1.40          & 69.58±2.31          & 86.38±0.61          \\
GCN                                                   & 55.41±3.47          & {\ul 44.07±1.95}    & 67.04±2.23           & 86.48±1.12          & 72.67±1.99          & 87.39±0.68          \\
H2GCN                                                 & 82.16±8.21          & 28.91±1.78          & 51.58±1.51           & {\ul 87.69±1.37}    & {\ul 75.95±2.18}    & 88.78±0.53          \\
GPR-GNN                                               & {\ul 84.59±4.37}    & 29.45±1.27          & \textbf{69.78±1.97 } & 86.70±1.03          & 75.12±1.98          & 87.38±0.63          \\
CPGNN-MLP                                             & 77.09±4.22          & 28.65±1.50          & 52.63±1.79           & 85.23±1.71          & 74.29±2.41          & 86.83±0.78          \\
CPGNN-Cheby                                           & 77.03±5.83          & 30.95±1.24          & 54.05±4.67           & 86.82±1.11          & 75.42±1.85          & {\ul 89.08±0.67}    \\
\textbf{BM-GCN(Ours)}                                 & \textbf{85.13±4.64} & \textbf{51.41±1.10} & {\ul 69.58±2.90}     & \textbf{87.99±1.29} & \textbf{76.13±1.92} & \textbf{90.25±0.75} \\ \bottomrule[1.5pt]
\end{tabular}
}
\caption{Node classification accuracy on six datasets. The best results are in bold and the second best results are underlined.}\label{table:Classification}
\end{table*}
\subsection{Experiment Setup}
\textbf{Datasets}. We conduct experiments on six real-world datasets with different homophily ratio. Among them, Cora, Citeseer and Pubmed~\cite{cora1,cora2,cora3} are three citation networks with high homophily ratios. Texas, Chameleon and Squirrel~\cite{chameleon} are three webpage datasets collected from university websites or Wikipedia, having low homophily ratios. The number of nodes, edges, classes, and features, as well as homophily ratios of the above datasets, are summarized in Table~\ref{table:datasets}.

\textbf{Baselines}. We compare our model with six baselines including a classic GCN~\cite{gcn}, an attribute-only based MLP, and four GNN-based SOTA methods aiming to analyze heterophilic graphs (i.e., H2GCN~\cite{h2gcn}, GPR-GNN~\cite{gprgnn}, CPGNN-MLP, and CPGNN-Cheby~\cite{cpgnn}). Among them, H2GCN aggregated more homophilic information by considering higher order neighborhoods. GPR-GNN tried to distill more valuable homophilic information by assigning signed weights to connected nodes. CPGNN propagated soft labels under the guidance of a compatibility matrix. CPGNN-MLP and CPGNN-Cheby are two variants of CPGNN with different soft label learning methods, i.e., MLP and GCN-Cheby~\cite{gcncheby}.

\textbf{Parameters.} For the baselines, we use their default parameter settings as they often lead to the best results. For our proposed method, we set the number of GCN layers $k$ to $2$ for Texas and $3$ for the other five datasets. We set the balance parameter of loss $\lambda$ to $0.5$, dropout ratio to $0.5$, learning rate to $0.001$, and weight decay to $0.0005$. We search on the enhancement factor $\alpha$ and self-loop coefficient $\beta$ from $0$ to $4$ for datasets. For all datasets, we use the same splits with Geom-GCN~\cite{geomgcn} and measure the performance of all models on the test sets over 10 random splits.

\subsection{Node Classification}
The results are shown in Table~\ref{table:Classification} where the best results are in bold fonts and the second-best results are underlined. The newly proposed method achieved the best performance on five of the six datasets while the second best performance on the remaining dataset. Considering the six datasets, our BM-GCN is on average 11.03\%, 7.91\%, 7.58\%, 4.58\%, 9.30\%, and 7.86\% more accurate than MLP, GCN, H2GCN, GPR-GNN, CPGNN-MLP, and CPGNN-Cheby respectively. To be specific, for the three datasets with heterophily (left in Table~\ref{table:Classification}), it is obvious that BM-GCN, H2GNN, CPGNN and GPR-GNN have higher mean accuracy than GCN and MLP. This indicates that special designs for heterophily are necessary when analyzing heterophilic graphs. Among these methods, our BM-GCN have the best performance, which demonstrates that our new idea of block-guided classified aggregation is sound and more useful for heterophilic graphs. For three datasets with homophily (right in Table~\ref{table:Classification}), the classic GCN has a good performance since it is designed under a strict homophily assumption and can be naturally applied to homophilic graphs. Other baselines can also achieve competitive performance. Despite these, our BM-GCN steadily outperforms all baselines, which demonstrates its generality.

\begin{figure*}[t]  
    \centering    
    
    \subfloat[H2GCN] 
    {
        \begin{minipage}[t]{0.19\textwidth}
            \centering          
            \includegraphics[width=0.98\textwidth]{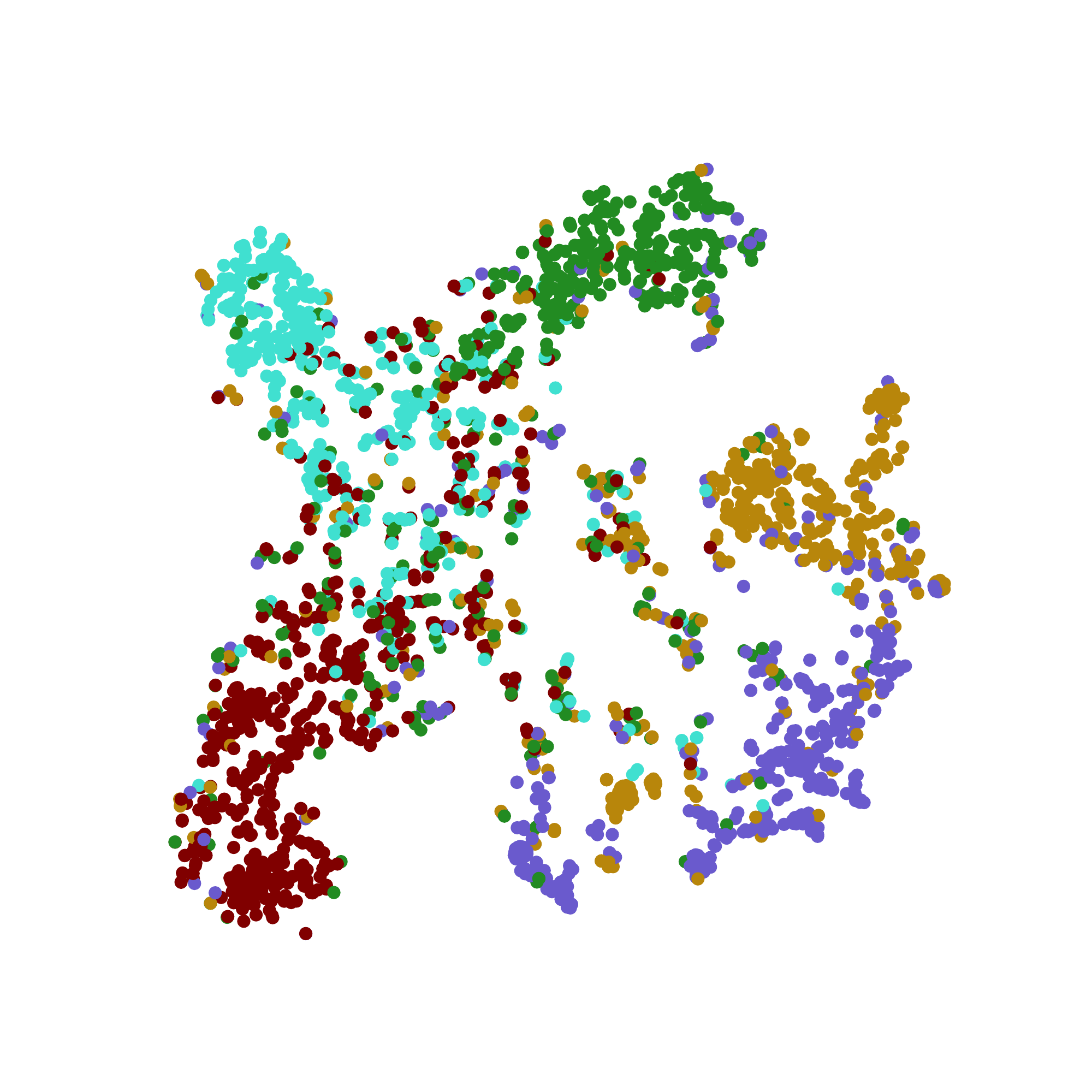}  
        \end{minipage}%
    }\hspace{5mm}
    \subfloat[CPGNN] 
    {
        \begin{minipage}[t]{0.19\textwidth}
            \centering      
            \includegraphics[width=0.98\textwidth]{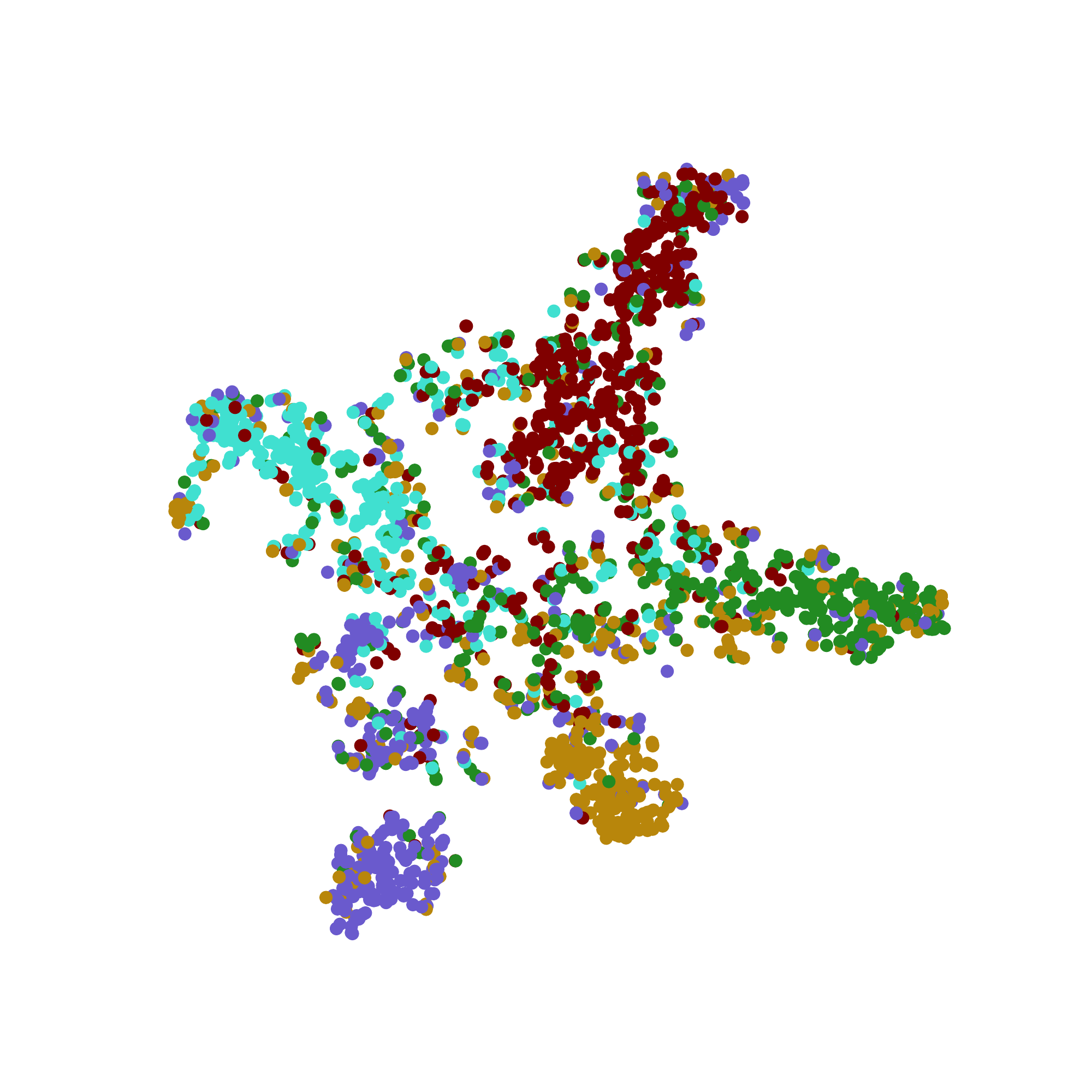}   
        \end{minipage}
    }\hspace{5mm}
    \subfloat[GPR-GNN] 
    {
        \begin{minipage}[t]{0.19\textwidth}
            \centering          
            \includegraphics[width=0.98\textwidth]{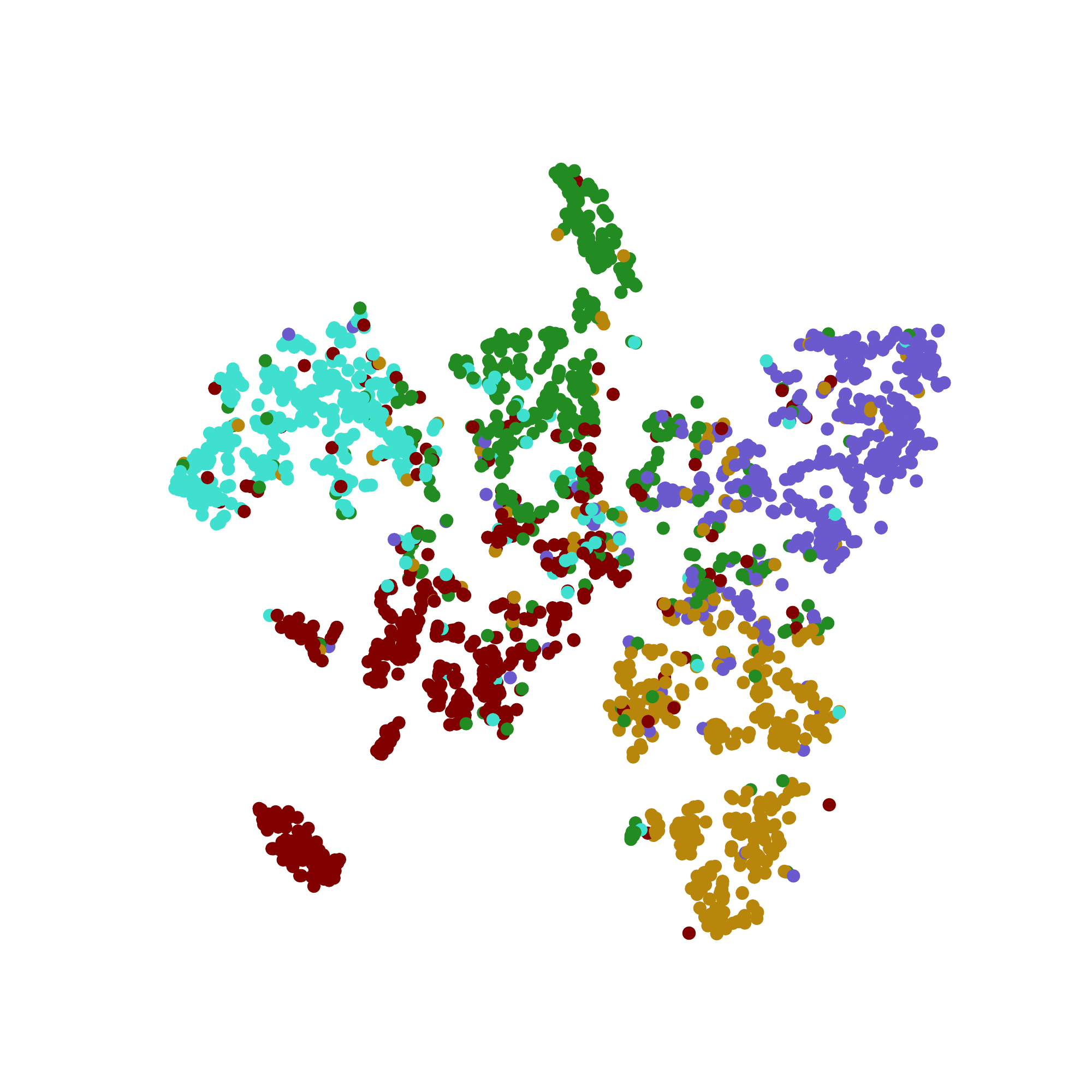}   
        \end{minipage}%
    }\hspace{5mm}
    \subfloat[BM-GCN] 
    {
        \begin{minipage}[t]{0.19\textwidth}
            \centering      
            \includegraphics[width=0.98\textwidth]{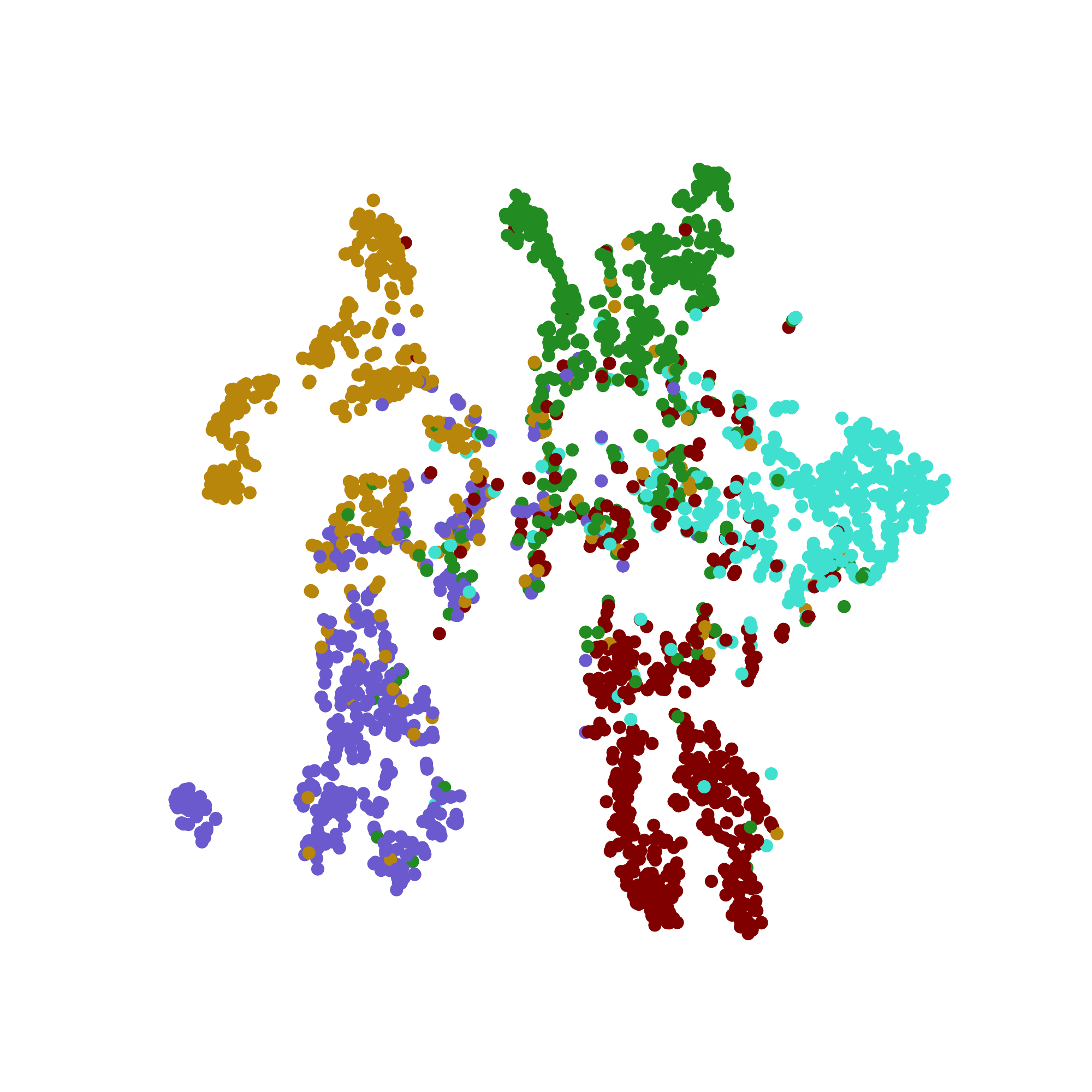}   
        \end{minipage}
    }
    
    \caption{Visualization results on Chameleon dataset. Different colors correspond to different ground truth classes.}\label{fig:visualization}
\end{figure*}

\subsection{Visualization}
In order to show the effectiveness of our proposed model in a more intuitive way, we further conduct visualization task on a heterophilic network Chameleon as an example. We extract the output embedding in the final layer of our BM-GCN as well as three SOTA baselines (CPGNN, GPR-GNN, and H2GCN) and present the learned embedding using t-SNE~\cite{tsne}. The result is shown in Fig.~\ref{fig:visualization}, in which nodes are colored by ground-truth labels.

From Fig.~\ref{fig:visualization} we can find that, the results of CPGNN, GPR-GNN and H2GCN are not so satisfactory as many nodes with different labels are mixed together. The performance of our proposed model is apparently the best, because the visualization of learned embeddings has a more compact structure, the highest intra-class tightness and the clearest boundaries among different classes. This further validate the effectiveness of our new idea of block-guided classified aggregation.

\subsection{Parameter Analysis}
We choose two datasets, i.e., a homophilic graph Cora and a heterophilic graph Chameleon to analyze some important hyper-parameters in BM-GCN, including the number of GCN layers $k$, the enhancement factor $\alpha$ in aggregating weight matrix $Q$, and the loss balancing parameter $\lambda$.

\textbf{Number of GCN layers $k$.} We test the node classification accuracy of BM-GCN with GCN layers $k$ varying from $1$ to $6$. The result are reported in Table~\ref{table-layer}. Results on the two datasets have the same trend. BM-GCN has a good and stable performance when $k=2,3,4$ and will have a performance degradation when $k$ is too large. This is a manifestation of the over-smoothing problem of GCN-based models. Even though, our BM-GCN can achieve competitive performance when $k=4$ while the classic GCN typically limits $k=2$. This is because BM-GCN uses matrix $Q$ to further optimized the topology so that noise information can be filtered to alleviate the over-smoothing problem.

\begin{table}[]
\resizebox{0.99\linewidth}{!}{
\begin{tabular}{lcccccc}
\toprule[1.5pt]
\multirow{2}{*}{\begin{tabular}[c]{@{}l@{}}Datasets/\\ Accuracy (\%)\end{tabular}} & \multicolumn{6}{c}{The number of graph convolutional layers $k$}  \\ \cmidrule(r){2-7}  
                                                                                                 & 1     & 2     & 3     & 4     & 5     & 6     \\ \midrule[0.6pt]
Cora                                                                                             & 81.27 & 86.92 & 87.99 & 87.30 & 60.10 & 40.16 \\
Chameleon                                                                                        & 58.71 & 67.74 & 69.58 & 65.00 & 49.28 & 34.69 \\ \bottomrule[1.5pt]
\end{tabular}
}
\caption{Node classification accuracy of BM-GCN with graph convolutional layers $k$ varying from 1 to 6.}\label{table-layer}
\end{table}

\begin{figure}[h]  
    \centering    
    
    \subfloat[Cora] 
    {
        \begin{minipage}[t]{0.21\textwidth}
            \centering          
            \includegraphics[width=0.98\textwidth]{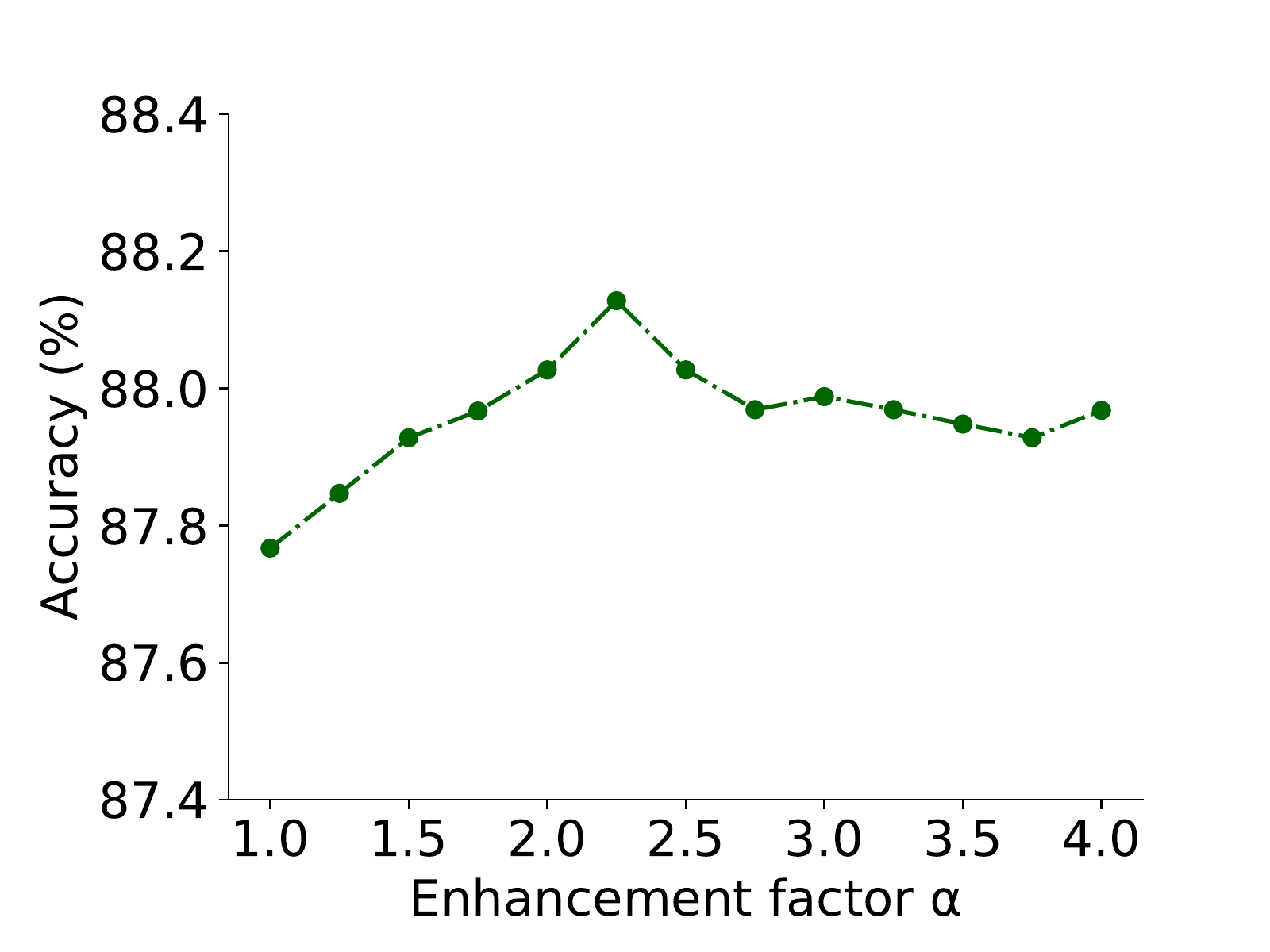}  
        \end{minipage}%
    }\hspace{5mm}
    \subfloat[Chameleon] 
    {
        \begin{minipage}[t]{0.21\textwidth}
            \centering      
            \includegraphics[width=0.98\textwidth]{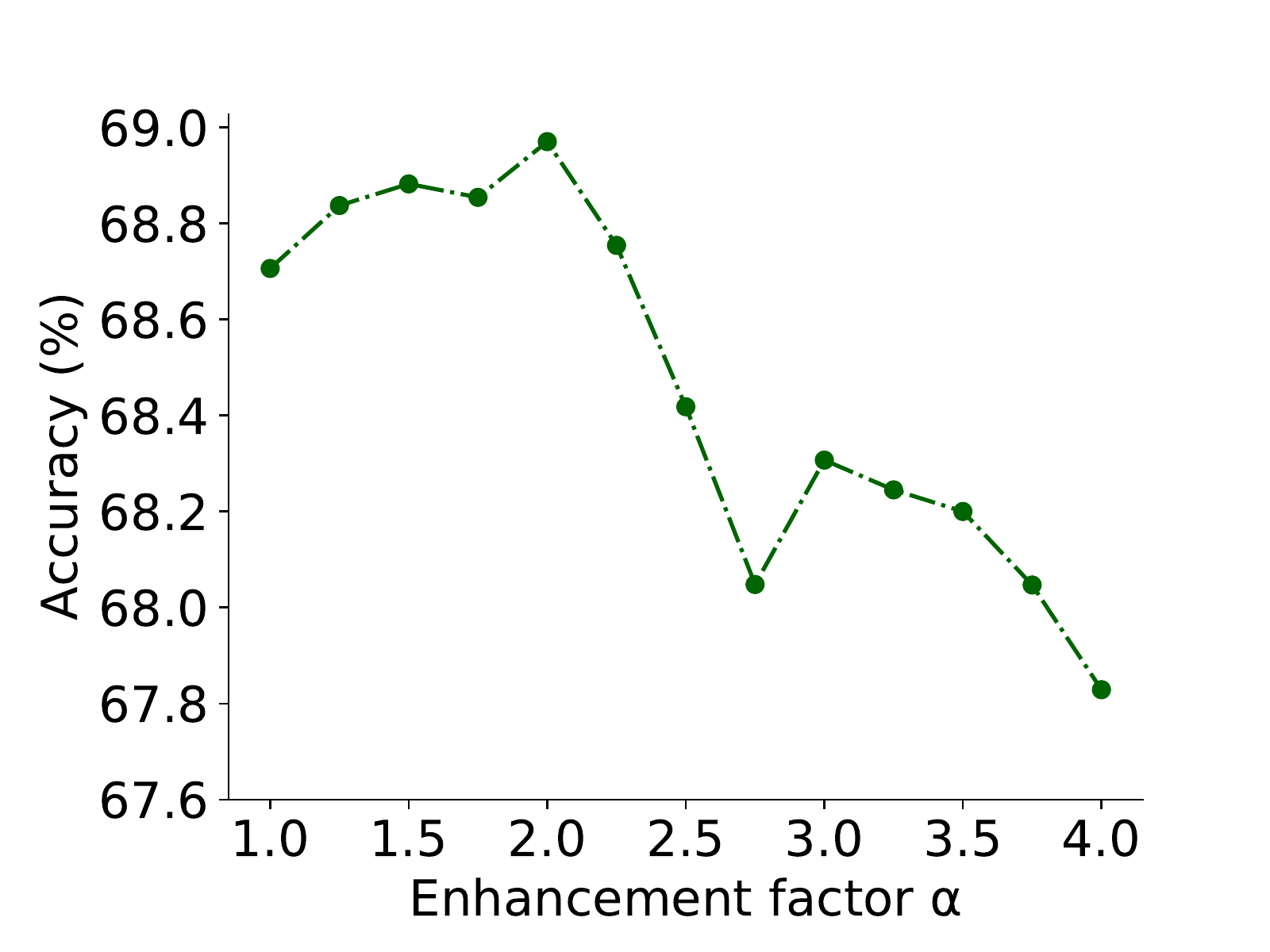}   
        \end{minipage}
    }
\caption{Parameter analysis of enhancement factor $\alpha$ in $Q$ on Cora and Chameleon datasets. We report the average node classification accuracy over 10 random splits.}\label{fig:enhance_Q}
\end{figure}

\textbf{Enhancement factor $\alpha $ in $Q$.} We test the node classification accuracy of BM-GCN with enhancement factor $\alpha$ varying from $1$ to $4$. The results are reported in Fig.~\ref{fig:enhance_Q}. BM-GCN achieves the best performance when $\alpha$ is around to $2$, which means that properly enhancing the diagonal elements of aggregating weight matrix $Q$ is effective since it can make nodes of the same class more closely connected. Even the performance will decrease when $\alpha$ is too large or too small, the performance changes is within 1\%. Empirically, it is feasible to set $\alpha$ around to $2$, which demonstrates the easy selectivity of the enhancement factor $\alpha$.

\begin{figure}[h]
    \centering 
    \subfloat[Cora] 
    {
        \begin{minipage}[t]{0.21\textwidth}
            \centering          
            \includegraphics[width=0.98\textwidth]{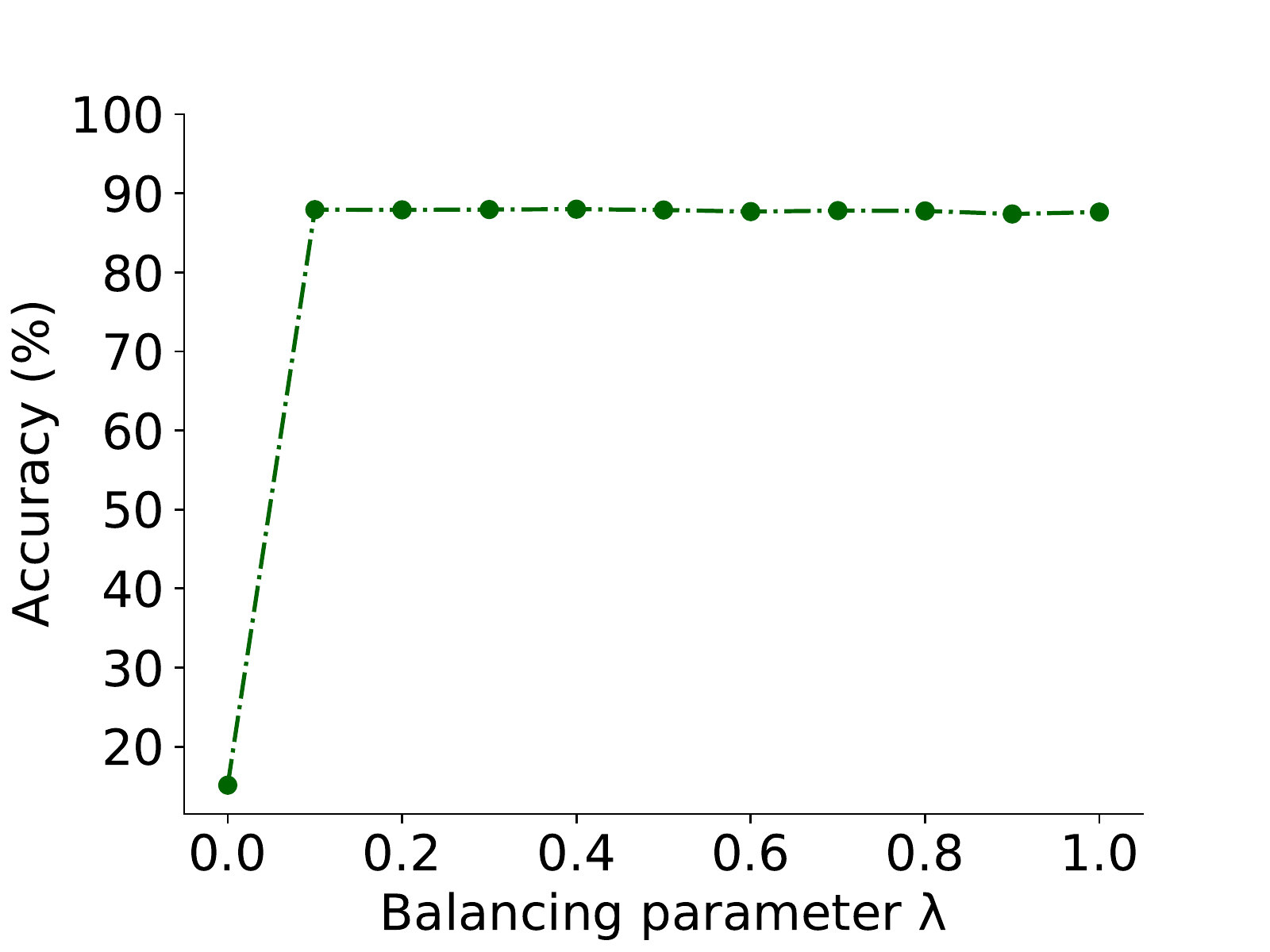}  
        \end{minipage}%
    }\hspace{5mm}
    \subfloat[Chameleon] 
    {
        \begin{minipage}[t]{0.21\textwidth}
            \centering      
            \includegraphics[width=0.98\textwidth]{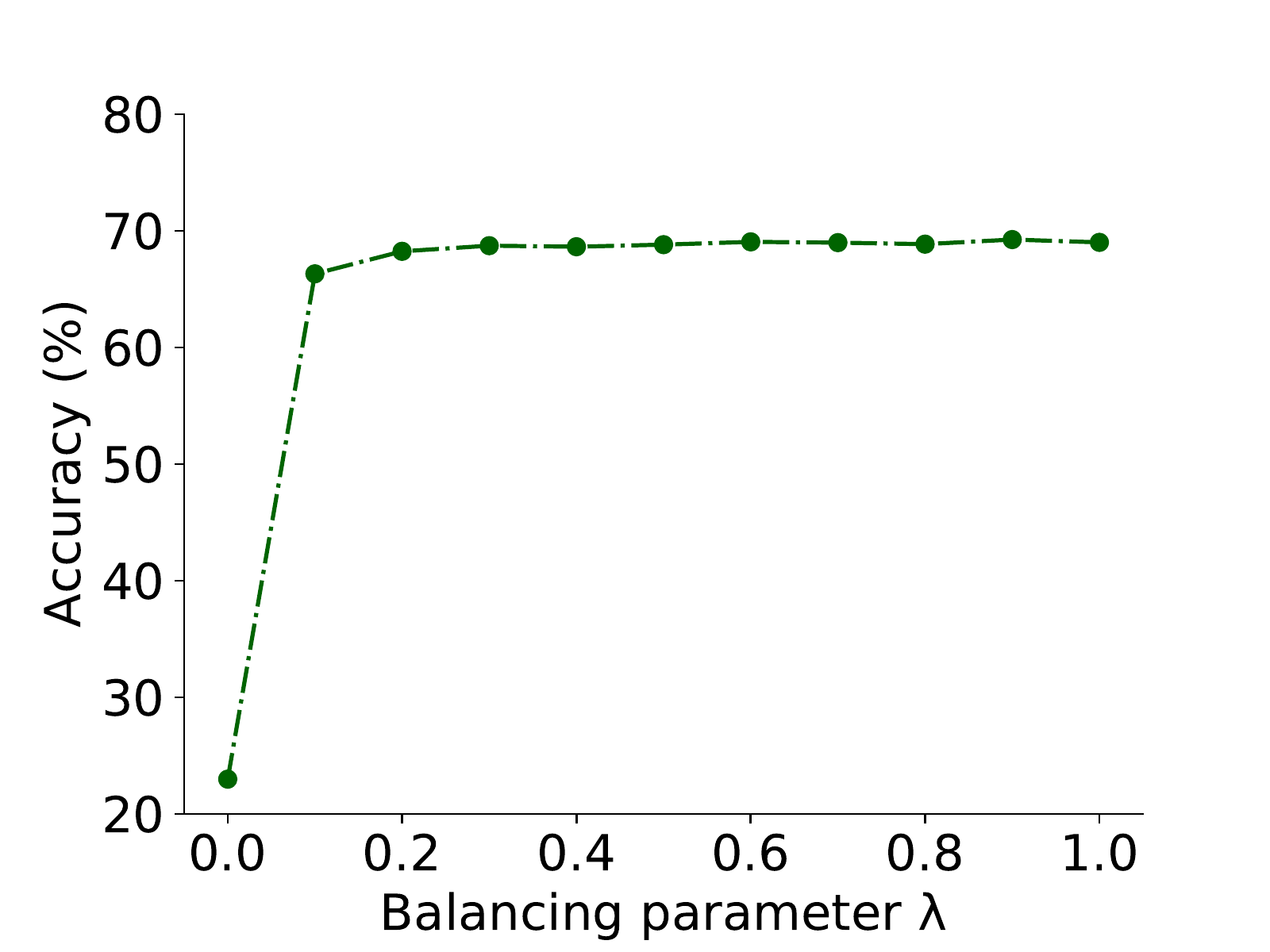}   
        \end{minipage}
    }
    \caption{Parameter analysis of loss balancing parameter $\lambda $ on Cora and Chameleon datasets. We report the average node classification accuracy over 10 random splits.}\label{fig:balance_lambda}
\end{figure}

\textbf{Loss balance parameter $\lambda $.} We test the node classification accuracy of BM-GCN with loss balancing parameter $\lambda$ varying from $0$ to $1$, and the result are reported in Fig.~\ref{fig:balance_lambda}. When $\lambda =0$, BM-GCN has a poor performance. This is because the parameters in GCN layers will not be optimized. When $\lambda >0$, BM-GCN has a stable and competitive performance. This demonstrates the robustness of loss balancing parameter in our proposed model.

\section{Conclusion}
In this paper, in order to develop a better graph convolutional operation universally suitable for both heterophily and homophily we propose a novel framework BM-GCN. The proposed BM-GCN introduces block modeling into graph convolutional operation in order to realize the principle of \emph{block guided classified aggregation}, which aggregates more information from neighbors of the same class while less from different classes. Specifically, BM-GCN consists of two main parts: one learns the similarity between classes according to the learned block matrix, and the other conducts graph convolutional operation with the guidance of block (class) similarity. Based on these two parts, we obtain a unified framework that achieves mutual optimization. We evaluate BM-GCN on both homogeneous and heterogeneous datasets. Experimental results on six real-world datasets verify the superiority of BM-GCN over the-state-of-art methods including those methods designed for heterophily.
% \newpage
\section{Acknowledgments}
This work is supported by the National Natural Science Foundation of China (61876128, 61832014, 61902278), and the Tianjin Municipal Science and Technology Project (19ZXZNGX00030).
\bibliography{aaai22}
\end{document}